\documentclass[11pt]{article}

\usepackage[preprint]{acl}

\usepackage{times}
\usepackage{latexsym}
\usepackage{amsmath,amssymb,amsfonts}
\usepackage{amsthm}
\usepackage{booktabs}
\usepackage{graphicx}
\usepackage{multirow}
\usepackage{enumitem}
\usepackage{xcolor}
\usepackage{algorithm}
\usepackage{algpseudocode}
\usepackage[T1]{fontenc}
\usepackage[utf8]{inputenc}
\usepackage{microtype}
\usepackage{xspace}

\newcommand{\sof}{\mathrm{softmax}}

\newtheorem{proposition}{Proposition}

\title{$\pi$-Attention: Online Efficient Sparse Transformers for Long-Context Modeling}

\author{Pike D. Liu$^{1}$, Chang Liu$^{2}$, Yanxuan Yu$^{3}$ \\
$^{1}$University of California, Los Angeles \\
$^{2}$University of Illinois Urbana-Champaign \\
$^{3}$Columbia University \\
\texttt{pikeliu@ucla.edu}, \texttt{cliu2645@gmail.com}, \texttt{yy3523@columbia.edu}}

\begin{document}
\maketitle

\begin{abstract}
Sparse attention is crucial in long-context Transformers, which restricts each token to a limited neighborhood and thereby reduces the quadratic cost of full self-attention. Local windows capture nearby context effectively, yet they induce a receptive-field bottleneck for dependencies beyond the window, limiting long-range modeling under moderate depth. In this paper, we propose $\pi$-Attention, an \emph{online efficient} sparse attention operator: as tokens arrive, each step maintains a streaming working set of local neighbors plus a $\pi$-indexed long-range fetch, fused by an adaptive prior under a shared softmax. Rather than materializing a global sparse mask in advance, $\pi$-Attention computes attention on the live working set with hierarchy-aware IO. We analyze causal reachability and minimum depth under this online rule, and show per-step cost remains $\mathcal{O}(k)$. Experiments on language modeling, Long Range Arena, and efficiency profiling---across 4K--32K context lengths---show consistent gains over local-window and other sparse baselines, approaching dense attention quality at linear cost.
\end{abstract}

\section{Introduction}
Transformer-based models \citep{vaswani2017attention} have delivered exceptional performances across widespread applications, including language processing \citep{devlin2018bert,brown2020language,raffel2019exploring}, computer vision \citep{dosovitskiy2020image,touvron2021training,liu2021swin}, and multimodal learning \citep{radford2021learning,lu2019vilbert}. Self-attention computes contextualized token representations by aggregating information from other positions according to query--key compatibility. Softmax normalization yields a probabilistic distribution over attended keys. Prior analyses observe that attention mass often concentrates on a small subset of tokens, motivating sparse patterns that discard many pairs without severely harming quality \citep{beltagy2020longformer,zaheer2020bigbird,child2019generating}. Structured sparse attention further offers predictable IO and scheduling properties on accelerators \citep{dao2022flashattention,liu2023ringattention}.

One primary limitation of \emph{sliding-window} (banded) sparse attention is the receptive-field bottleneck. When each token attends only within radius $k$, information expands by at most $k$ positions per layer, so depth-$L$ models cover only $\mathcal{O}(kL)$ span \citep{bengio1994learning,hua2022transformer}. Long-document and retrieval settings amplify this gap. Alternatives include dilated/strided patterns \citep{child2019generating,li2019enhancing}, global tokens \citep{beltagy2020longformer}, random--global hybrids \citep{zaheer2020bigbird}, and routing sparsity \citep{roy2021efficient}. These help coverage but often introduce irregular communication, content-dependent graphs, or weaker hardware locality.

To address this limitation, we propose $\pi$-Attention as an \emph{online efficient compute} paradigm for sparse attention. At each decoding/training step $t$, the operator streams a working set $\mathcal{W}_t$ of local neighbors together with a $\pi$-indexed long-range partner, applies adaptive fusion on the fly, and never requires a prebuilt full-sequence mask. The parameter $\pi$ controls the online long-range fetch schedule and the hardware-friendly gather period; it is not a claim of binary-lifting multi-scale hops. We analyze reachability under this streaming rule and give a hierarchical online cost model. The block drops into standard Transformers with minor changes.
\begin{enumerate}[leftmargin=*,itemsep=1pt,topsep=2pt]
\item We propose $\pi$-Attention, an online efficient sparse operator that streams local windows with a $\pi$-indexed long-range fetch and adaptive fusion under a shared softmax.
\item We analyze causal reachability and minimum depth for the online working set, and formulate a multi-level hardware cost $C_{\mathrm{online}}(n)$ for efficient streaming compute.
\item We evaluate against dense attention, local-window attention, and representative sparse baselines on language modeling, LRA, and efficiency metrics, with ablations over skips and fusion.
\end{enumerate}

\section{Related Works}
\paragraph{Transformer Attention.}
The Transformer model, introduced by \citet{vaswani2017attention}, revolutionized NLP with its self-attention mechanism. Unlike previous sequence models such as RNNs and LSTMs, Transformer does not rely on recurrent structures and instead uses self-attention to depict relationships between input tokens in parallel. Self-attention, also known as scaled dot-product attention, computes attention scores between input tokens using the query ($Q$), key ($K$), and value ($V$) vectors:
\begin{equation}
\mathrm{Attention}(Q,K,V)=\sof\!\left(\frac{QK^\top}{\sqrt{d_k}}\right)V,
\end{equation}
where $d_k$ is the dimension of the key vectors \citep{vaswani2017attention}. There are also linearized attention methods, such as Linformer \citep{wang2020linformer} and Performer \citep{choromanski2021rethinking}, approximating the softmax attention function using low-rank or kernel approximations, reducing the computational complexity from $\mathcal{O}(n^2)$ to $\mathcal{O}(n)$. Another approach to reduce computational complexity is through sparse attention, where only a subset of attention scores are computed. For example, Longformer \citep{beltagy2020longformer} uses a combination of local windowed attention and global attention, reducing the attention complexity to $\mathcal{O}(n)$ for sequences of length $n$.

\paragraph{Sparse Attention and Receptive Fields.}
BigBird \citep{zaheer2020bigbird} combines random, local, and global attention; Longformer \citep{beltagy2020longformer} uses sliding windows with global tokens; Sparse Transformers \citep{child2019generating} introduce strided/dilated patterns; LogSparse-style designs use exponentially growing intervals for long-range time series \citep{li2019enhancing}; Routing Transformer \citep{roy2021efficient} learns content-based sparsity; Reformer \citep{kitaev2020reformer} uses LSH. Separately, Ring Attention \citep{liu2023ringattention} is a \emph{distributed exact dense} attention execution strategy via blockwise ring communication---not a local sparse mask---and FlashAttention \citep{dao2022flashattention} accelerates exact attention with IO awareness. Limited receptive fields remain a classic obstacle for deep sequence models \citep{bengio1994learning}. Our method is related to local+strided sparse attention, but is positioned as \emph{online efficient compute}: (i)~a streaming working set with a $\pi$-indexed long-range fetch and learned local/skip prior under one softmax, (ii)~causal reachability analysis for that online rule, and (iii)~a hierarchy-aware cost model for live IO. We do \emph{not} claim Ring Attention semantics for our LocalWindow baseline.

\section{Method}
\subsection{Softmax Attention Mechanism}
In the attention mechanism, the weight $\alpha_{ij}$ represents the attention score between token $i$ (the query) and token $j$ (the key). This score quantifies the relative importance of token $j$ to token $i$, among all tokens in the input sequence. It is formulated as
\begin{equation}
\alpha_{ij}=\sof\!\left(\frac{q_i^\top k_j}{\sqrt{d_k}}\right)
=\frac{\exp\!\big(q_i^\top k_j/\sqrt{d_k}\big)}
{\sum_{j'}\exp\!\big(q_i^\top k_{j'}/\sqrt{d_k}\big)},
\label{eq:alpha}
\end{equation}
where $q_i$ and $k_j$ are the query and key vectors for tokens $i$ and $j$, respectively, and $d_k$ is a scaling factor based on the dimensionality of the keys \citep{vaswani2017attention}. The softmax function ensures that the resulting attention scores $\alpha_{ij}$ are normalized and can be interpreted as probabilities, summing to one over all tokens $j$ for a given query token $i$.

The final output of the attention mechanism for each query token $i$ is then calculated as a weighted sum of the values $v_j$ corresponding to each token $j$ in the sequence, with the weight determined by the attention scores $\alpha_{ij}$. The output of the attention mechanism for token $i$ is defined as
\begin{equation}
\mathrm{Attention}_i(Q,K,V)=\sum_{j}\alpha_{ij}v_j,
\label{eq:attn}
\end{equation}
where $Q$, $K$, and $V$ are matrices representing all queries, keys, and values for a given sequence. This approach allows the model to focus selectively on parts of the sequence that contribute meaningfully to the current query position \citep{bahdanau2015neural}.

\subsection{Local Sparse Attention}
Training long-context Transformer models involves restricting attention to a sparse neighborhood to avoid quadratic cost. We denote the pre-softmax attention scores as
\begin{equation}
z_{i,j}=\frac{q_i^\top k_j}{\sqrt{d_k}},
\label{eq:z}
\end{equation}
and define the local-window neighborhood of token $i$ as
\begin{equation}
\mathcal{N}_r(i)=
\begin{cases}
\{j:\max(0,i-k)\le j\le i\}, & \text{causal},\\
\{j:\,|j-i|\le k\}, & \text{bidirectional},
\end{cases}
\label{eq:Nr}
\end{equation}
where $k$ is the local window size and the causal case includes the self-position $j=i$. The corresponding local sparse attention replaces the full sum in Equation~\eqref{eq:attn} by a sum over $j\in\mathcal{N}_r(i)$ only. This formulation preserves locality and yields linear complexity $\mathcal{O}(nk)$, but also makes long-range modeling challenging, as explored in the next section.

\subsection{Receptive-Field Bottleneck in Local Sparse Attention}
One notable issue with local sparse attention is the receptive-field bottleneck, especially when dependencies lie far beyond the local window. When each layer expands reachability by at most $k$ positions, the gradients and representations from distant tokens can become excessively weak or even unreachable, slowing down or even preventing long-range learning. This is particularly problematic in deeper models where multiple layers of attention are stacked.

The bottleneck arises from the form of layered neighborhood composition. We examine the two cases: Consider the token $i$, and let $\mathcal{R}_L(i)$ denote the set of positions that can reach $i$ after $L$ layers under causal local attention. In the extreme case where a dependency at $i-d$ satisfies $d>kL$, we have
\begin{equation}
i-d\notin\mathcal{R}_L(i),
\label{eq:unreachable}
\end{equation}
so the corresponding information cannot affect the prediction at $i$. Moreover, in a milder case where $d\le kL$ but $d$ is large, many intermediate hops are required:
\begin{equation}
|\mathcal{R}_L(i)|\le kL,
\label{eq:rf_local}
\end{equation}
which means that intermediate layers must dedicate capacity to transporting information rather than transforming it, resulting in suboptimal training performance.

In summary, when the extreme case arises where a dependency lies outside the local receptive field, local sparse attention suffers from complete information blocking for that dependency, leading to weak long-range modeling. In the milder case, where the dependency is reachable only through many local hops, the associated pathway still dilutes, causing suboptimal training performance. The extreme case frequently occurs in long-document and retrieval settings due to large token distances. In the following section, we introduce $\pi$-Attention as an online efficient sparse operator over a streaming working set, then analyze reachability and present fusion variants.

\subsection{\texorpdfstring{$\pi$-Attention}{pi-Attention}}
To address limited receptive-field growth \emph{and} quadratic/global-mask cost, we propose $\pi$-Attention as an online efficient compute rule. At position $i$ (or streaming step $t{=}i$), attention is evaluated only on the live union working set
\begin{equation}
\mathcal{U}(i)=\mathcal{N}_r(i)\cup\mathcal{N}_{\pi}(i),
\label{eq:U}
\end{equation}
where the online long-range partner is
\begin{equation}
\mathcal{N}_{\pi}(i)=\{i-\pi\}
\label{eq:Npi}
\end{equation}
under causal decoding (bidirectional tasks may also fetch $i+\pi$). Here $\pi$ is a positive integer that parameterizes the \emph{online long-range fetch period}---not the mathematical constant and not a precomputed multi-scale mask. Local context stays in the sliding window; the $\pi$-fetch brings a distant key/value into the working set when needed, enabling efficient streaming compute.

\begin{figure}[t]
\centering
\includegraphics[width=\columnwidth]{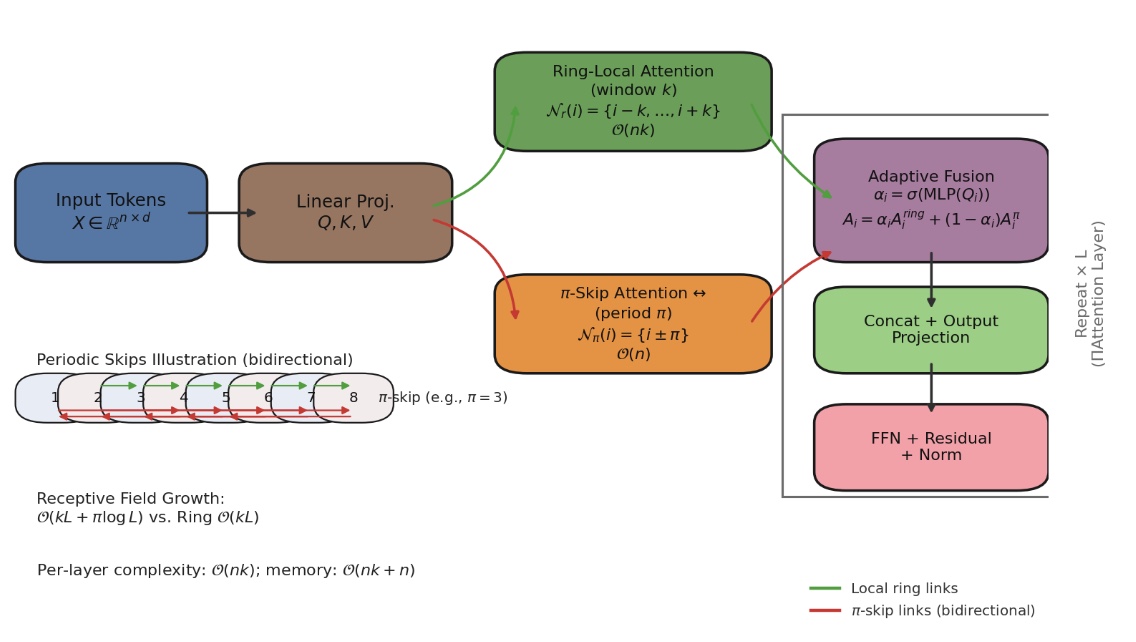}
\caption{Illustration of online $\pi$-Attention: streaming local-window neighbors, a $\pi$-indexed long-range fetch, and adaptive fusion over the live working set.}
\label{fig:concept}
\end{figure}

We then compute attention over the union set. Let
\begin{equation}
\alpha_{i,j}=\sof(z_{i,\mathcal{U}(i)})_j,
\label{eq:alpha_u}
\end{equation}
denote the softmax probability of neighbor $j\in\mathcal{U}(i)$. The basic $\pi$-Attention output is
\begin{equation}
\beta_{i,j}=\alpha_{i,j},\qquad
\mathrm{Attention}_i=\sum_{j\in\mathcal{U}(i)}\beta_{i,j}v_j,
\label{eq:beta}
\end{equation}
where $z_{i,\mathcal{U}(i)}$ is the pre-softmax score vector restricted to $\mathcal{U}(i)$. This formulation preserves a single softmax over the live working set and keeps per-step complexity linear in the window size $k$.

\paragraph{Reachability analysis of $\pi$-Attention.}
Under causal $\pi$-Attention, each layer may take a backward local hop of length at most $k$ (including $0$) and/or one online $\pi$-fetch of length $\pi$. After $L$ layers, the positions that can reach query $i$ are
\begin{equation}
\mathcal{R}_L(i)=\bigl\{i-a-\pi b:\ 0\le b\le L,\ 0\le a\le k(L-b)\bigr\},
\label{eq:reachable}
\end{equation}
so the maximum span admits the tight linear upper bound
\begin{equation}
R(L)\le L(k+\pi),
\label{eq:rf_pi}
\end{equation}
whereas purely local windows satisfy only $R(L)\le kL$ (Equation~\eqref{eq:rf_local}). Importantly, a single online $\pi$-fetch per layer does \emph{not} induce binary lifting: there is no mechanism that creates hops of length $\pi,2\pi,4\pi,\ldots$ within one step, so logarithmic claims of the form $\pi\lceil\log_2 L\rceil$ are incorrect for this operator.

\paragraph{Minimum depth for distance $d$.}
For a target lag $d\ge 0$, the minimum number of layers needed is
\begin{equation}
L_{\min}(d)=\min_{b\ge 0}\left[b+\left\lceil\frac{\max(0,d-b\pi)}{k}\right\rceil\right],
\label{eq:lmin}
\end{equation}
with the convention $\lceil 0/k\rceil=0$. Compared with the local-only depth $\lceil d/k\rceil$, choosing $b\approx d/\pi$ reduces hop count whenever $\pi>k$. Coverage holes can still arise if $\gcd$ constraints and truncation interact poorly with $\pi$; we therefore treat $\pi$ as a tuned online-fetch period and validate it empirically.

\paragraph{Implications.}
When $d>kL$, local windows cannot connect the dependency at all, while $\pi$-Attention remains feasible once $L(k+\pi)\ge d$. When $d\le kL$, skips can still shorten the path via Equation~\eqref{eq:lmin}. The gate then decides whether to emphasize the skip edge for a given query. This yields a simple online alternative to random or multi-scale sparse masks, at the cost of one extra live candidate per step.

\paragraph{Comparison with local-window attention.}
Local-window attention confines reachability to $\mathcal{O}(kL)$. Replacing $\mathcal{N}_r(i)$ by the online working set $\mathcal{U}(i)$ adds one $\pi$-indexed long-range fetch per layer and raises the span bound to $\mathcal{O}(L(k+\pi))$. This does not recover dense attention, but it mitigates the local bottleneck while keeping gathers streaming-friendly.

\subsection{Variants of $\pi$-Attention}
In this section, we further develop some variants of $\pi$-Attention, by utilizing fusion techniques on the local and skip neighbors to further stabilize the training process.

\paragraph{Variant 1: separate softmax.}
A notable potential inconsistency arises when local and skip neighbors are scored independently and then mixed after separate normalization, which prevents cross-branch scores from living in a common simplex. To address this issue, we keep a single softmax over $\mathcal{U}(i)$ as in Equation~\eqref{eq:beta}. When separate softmaxes are used for diagnosis, we write
\begin{equation}
\gamma_{i,j}=
\begin{cases}
\sof(z_{i,\mathcal{N}_r(i)})_j, & j\in\mathcal{N}_r(i),\\
\sof(z_{i,\mathcal{N}_{\pi}(i)})_j, & j\in\mathcal{N}_{\pi}(i),
\end{cases}
\label{eq:gamma}
\end{equation}
and observe that $\gamma_{i,j}$ is generally not comparable across the two branches.

\paragraph{Variant 2: fixed fusion.}
Building on the union softmax, a more controlled technique is fixed fusion. We introduce a scalar prior $\alpha\in(0,1)$ and redefine the logits as
\begin{equation}
\delta_{i,j}=z_{i,j}+\log\omega_{i,j},
\label{eq:delta}
\end{equation}
where $\omega_{i,j}=\alpha$ if $j\in\mathcal{N}_r(i)$ and $\omega_{i,j}=1-\alpha$ if $j\in\mathcal{N}_{\pi}(i)$. The attention scores are then
\begin{equation}
\hat{\alpha}_{i,j}=\sof(\delta_{i,\mathcal{U}(i)})_j.
\label{eq:alphahat}
\end{equation}
This normalization ensures that the local/skip preference lies within a bounded prior, resulting in a stable mixing effect across different input distributions.

\paragraph{Variant 3: adaptive fusion.}
To make it easier for model optimization, we replace the fixed prior by a content-dependent gate
\begin{equation}
\alpha_i=\sigma\!\big(\mathrm{MLP}(q_i)\big)\in(0,1),
\label{eq:gate}
\end{equation}
and set
\begin{equation}
\omega_{i,j}=
\begin{cases}
\alpha_i, & j\in\mathcal{N}_r(i),\\
1-\alpha_i, & j\in\mathcal{N}_{\pi}(i).
\end{cases}
\label{eq:omega}
\end{equation}
The fused logits and attention then become
\begin{align}
\delta_{i,j}&=z_{i,j}+\log\omega_{i,j},
\label{eq:delta_adapt}\\
\hat{\alpha}_{i,j}&=\frac{\omega_{i,j}\exp(z_{i,j})}{\sum_{t\in\mathcal{U}(i)}\omega_{i,t}\exp(z_{i,t})},
\label{eq:alphahat_adapt}\\
\mathrm{Attention}_i&=\sum_{j\in\mathcal{U}(i)}\hat{\alpha}_{i,j}v_j.
\label{eq:attn_adapt}
\end{align}
When $\alpha_i$ approaches $1$, adaptive fusion degrades to local-window attention; when $\alpha_i$ approaches $0$, skip neighbors receive a stronger prior. For numerical stability we use $\tilde{\alpha}_i=(1-2\varepsilon)\alpha_i+\varepsilon$ with $\varepsilon=10^{-4}$. By default, we use Variant~3.

\paragraph{Gradient analysis of adaptive fusion.}
Let us evaluate the gradient of $\hat{\alpha}_{i,j}$ with respect to the pre-softmax score $z_{i,j'}$. Differentiating Equation~\eqref{eq:alphahat_adapt} yields the standard softmax Jacobian on the fused logits,
\begin{equation}
\frac{\partial\hat{\alpha}_{i,j}}{\partial z_{i,j}}=\hat{\alpha}_{i,j}\big(1-\hat{\alpha}_{i,j}\big),\qquad
\frac{\partial\hat{\alpha}_{i,j}}{\partial z_{i,j'}}=-\hat{\alpha}_{i,j}\hat{\alpha}_{i,j'}\ (j'\neq j).
\label{eq:jac}
\end{equation}
Moreover, differentiating with respect to the gate gives
\begin{equation}
\frac{\partial\delta_{i,j}}{\partial\alpha_i}=
\begin{cases}
1/\alpha_i, & j\in\mathcal{N}_r(i),\\
-1/(1-\alpha_i), & j\in\mathcal{N}_{\pi}(i),
\end{cases}
\label{eq:ddelta_da}
\end{equation}
and therefore
\begin{equation}
\frac{\partial\hat{\alpha}_{i,j}}{\partial\alpha_i}
=\hat{\alpha}_{i,j}\Biggl(
\frac{\partial\delta_{i,j}}{\partial\alpha_i}
-\sum_{t\in\mathcal{U}(i)}\hat{\alpha}_{i,t}\frac{\partial\delta_{i,t}}{\partial\alpha_i}
\Biggr).
\label{eq:dalpha_da}
\end{equation}
Equation~\eqref{eq:dalpha_da} shows that the gate provides a self-adjusting pathway: increasing $\alpha_i$ re-masses probability toward local neighbors, while decreasing $\alpha_i$ re-masses toward skips, without breaking the single-softmax ranking over $\mathcal{U}(i)$.

\paragraph{Implications for fusion stability.}
According to Equations~\eqref{eq:alphahat_adapt}--\eqref{eq:dalpha_da}, considering the extreme case where $\alpha_i\approx 1$, the skip prior vanishes and the mechanism reduces to local sparse attention. Moreover, for tokens where long-range evidence is needed, a smaller $\alpha_i$ amplifies skip logits before softmax, as demonstrated in Equation~\eqref{eq:delta_adapt}. Therefore, Variant~3 significantly enhances controllability of local/skip trade-offs through the dynamic gate, while preserving a valid probability simplex on $\mathcal{U}(i)$.

\paragraph{Comparison with separate softmax.}
In Variant~1, local and skip branches are normalized independently, so their probabilities are not comparable on a shared simplex. A shared softmax on prior-adjusted logits $\delta_{i,j}$ yields a jointly normalized distribution over local and skip candidates, allowing prior-adjusted scores to be compared within one probability simplex. Note that $\log\omega_{i,j}$ \emph{does} change rankings relative to raw $z_{i,j}$; what is preserved is joint normalization of the fused logits, not the ranking of $z$.

\paragraph{Complexity.}
Let $|\mathcal{U}(i)|=m=\mathcal{O}(k)$. The per-layer cost is
\begin{equation}
\mathrm{Cost}=
\underbrace{\mathcal{O}(nmd_h)}_{\text{union scores}}
+\underbrace{\mathcal{O}(nd_h)}_{\text{skip gather}}
+\underbrace{\mathcal{O}(nd_h)}_{\text{gate MLP}}
=\mathcal{O}(nkd_h).
\label{eq:cost}
\end{equation}

\paragraph{Hardware-aware online execution.}
We view $\pi$-Attention as an \emph{online} streaming operator over the arriving token stream, rather than a prebuilt sparse mask over the full sequence. At step $t$, only the working set
\begin{equation}
\mathcal{W}_t=\mathcal{N}_r(t)\cup\mathcal{N}_{\pi}(t)
\label{eq:online_ws}
\end{equation}
must be resident; older tokens outside $\mathcal{W}_t$ can remain in slower memory or on remote devices until a later hop brings them into range. This yields an IO-aware recurrence in the spirit of FlashAttention \citep{dao2022flashattention} and Ring Attention \citep{liu2023ringattention}: local scores are accumulated tile-by-tile in on-chip buffers, the gate and union softmax of Equations~\eqref{eq:delta_adapt}--\eqref{eq:alphahat_adapt} are applied on the fly, and skip keys/values are fetched when the current tile needs them---not because a global schedule was fixed in advance.

To make the hierarchy explicit, let $\gamma_{\mathrm{reg}}$, $\gamma_{\mathrm{sram}}$, $\gamma_{\mathrm{hbm}}$, and $\gamma_{\mathrm{net}}$ denote effective throughputs of registers, shared memory / SRAM, device HBM, and inter-device interconnect, and let $b_{\ell}$ be the bytes touched at level $\ell\in\{\mathrm{reg},\mathrm{sram},\mathrm{hbm},\mathrm{net}\}$ while processing a length-$n$ stream. The hardware-aware online cost is
\begin{equation}
\begin{aligned}
&\resizebox{0.90\columnwidth}{!}{$\displaystyle
C_{\mathrm{online}}(n)
=\dfrac{c_{\mathrm{tc}}n|\mathcal{W}|d_h}{\gamma_{\mathrm{tc}}}
+\sum_{\ell\in\mathcal{L}}
\dfrac{b_{\ell}(n;k,\pi,\alpha)}{\gamma_{\ell}},$}
\\
\mathcal{L}&=\{\mathrm{reg},\mathrm{sram},\mathrm{hbm},\mathrm{net}\},
\\
b_{\mathrm{reg}}&=\Theta(n|\mathcal{W}|d_h),\ 
b_{\mathrm{sram}}=\Theta(n_{\mathrm{tile}}|\mathcal{W}|d_h),
\\
b_{\mathrm{hbm}}&=\Theta(n(k{+}1)d_h),\ 
b_{\mathrm{net}}=\Theta(n_{\mathrm{remote}}(\pi)\,d_h).
\end{aligned}
\label{eq:online_cost}
\end{equation}
where $\gamma_{\mathrm{tc}}$ is tensor-core throughput, $|\mathcal{W}|=\mathcal{O}(k)$ is the online working-set size, $n_{\mathrm{tile}}$ is the SRAM tile width chosen at runtime from available shared memory, and $n_{\mathrm{remote}}(\pi)$ counts only those skip partners that currently reside off-device. Because $b_{\mathrm{hbm}}$ and $b_{\mathrm{net}}$ grow with the stream rather than with $n^2$, and because $\alpha_t$ is computed from the live query $q_t$, the algorithm adapts fusion and memory movement as tokens arrive. Empirically, this online hierarchy keeps skip/fusion overhead small relative to local compute while improving long-range quality (Section~\ref{sec:exp}). A reference implementation is given in Appendix~\ref{app:impl}.

\begin{proposition}[Causal reachability under online $\pi$-fetch]
\label{prop:rf}
Under causal $\pi$-Attention with local radius $k$ and online fetch period $\pi$, if each layer allows a backward local displacement in $\{0,\ldots,k\}$ and at most one $\pi$-fetch, then after $L$ layers the reachable set is exactly Equation~\eqref{eq:reachable} (up to boundary truncation), and
\begin{equation}
R(L)\le L(k+\pi).
\label{eq:prop}
\end{equation}
Moreover, the minimum depth to cover lag $d$ is given by Equation~\eqref{eq:lmin}.
\end{proposition}
\begin{proof}[Proof sketch]
Each layer contributes one summand of the form $a_\ell+\pi b_\ell$ with $a_\ell\in\{0,\ldots,k\}$ and $b_\ell\in\{0,1\}$. Summing over $L$ layers yields $a=\sum a_\ell\le k(L-b)$ and $b=\sum b_\ell\le L$, which is Equation~\eqref{eq:reachable}. The span bound follows by maximizing $a+\pi b$. Minimizing $b+\lceil\max(0,d-b\pi)/k\rceil$ over integers $b\ge 0$ yields Equation~\eqref{eq:lmin}.
\end{proof}

\section{Experiment}
\label{sec:exp}
\paragraph{Datasets.}
We use WikiText-103 and PG-19 for language modeling, LRA (ListOps, Retrieval, Pathfinder; accuracy) for long-range reasoning, and MSCOCO and Flickr30K for vision-language retrieval. The training is conducted with batch size 256 and sequence lengths from 4K to 32K depending on the task. The models are trained for 200K iterations. Throughout the training process, we monitor both validation perplexity and efficiency. We also evaluate our methods in ablations over skips and fusion.

\paragraph{Experiment Setting.}
We begin by conducting experiments on WikiText-103 and PG-19, evaluating model performance against dense and sparse baselines. Next, we validate our method on LRA and vision-language retrieval. Following this, we analyze efficiency and different variants. Subsequently, we further validate the method with visualizations. The experiment setting details are presented in Appendix~\ref{app:exp}. By default, we use online Variant~3 with $k=4$ and fetch period $\pi=16$. Main quality tables report mean$\pm$std over three seeds.

\subsection{Compare with Baseline Performance}
\begin{table}[t]
\centering
\caption{Perplexity (mean$\pm$std over 3 seeds) on WikiText-103 and PG-19. LocalWindow is sliding-window sparse attention (not Ring Attention).}
\label{tab:lm}
\setlength{\tabcolsep}{4pt}
\small
\begin{tabular}{lcc}
\toprule
Method & WikiText-103 & PG-19 \\
\midrule
Full Attention & $18.3{\pm}0.05$ & $12.7{\pm}0.04$ \\
LocalWindow & $20.1{\pm}0.12$ & $14.2{\pm}0.09$ \\
BigBird & $19.8{\pm}0.10$ & $13.9{\pm}0.08$ \\
Longformer & $19.5{\pm}0.09$ & $13.6{\pm}0.07$ \\
Reformer & $20.5{\pm}0.14$ & $14.8{\pm}0.11$ \\
\midrule
$\pi$-Attention & $\mathbf{18.4{\pm}0.06}$ & $\mathbf{12.9{\pm}0.05}$ \\
\bottomrule
\end{tabular}
\end{table}

\noindent\textbf{$\pi$-Attention improves language modeling over local-window baselines.}
The results in Table~\ref{tab:lm} highlight the effectiveness of $\pi$-Attention in improving perplexity relative to local-window attention and other sparse methods. Without $\pi$-Attention, LocalWindow achieves 20.1 and 14.2 on WikiText-103 and PG-19. With $\pi$-Attention, these values drop to 18.4 and 12.9, showcasing its contribution. Longformer and BigBird remain between local-window attention and $\pi$-Attention. These results suggest that online $\pi$-Attention can improve language modeling over the evaluated local-window setting.

\noindent\textbf{$\pi$-Attention approaches dense attention under linear cost.}
Among all tested configurations, $\pi$-Attention yields the lowest perplexity among linear-complexity methods, nearly matching full attention (18.3 / 12.7). This superiority is consistent across both corpora, confirming that $\pi$-Attention is effective for reducing perplexity in language modeling tasks.

\noindent\textbf{The proposed $\pi$-Attention improves both short and long-document modeling.}
The analysis indicates that $\pi$-Attention enhances performance on WikiText-103 and the longer-form PG-19 corpus, demonstrating its ability to handle both moderate and long-range dependencies effectively. The reductions in perplexity are kept on PG-19, suggesting that $\pi$-Attention still provides better optimization for long contexts. This capability is critical for modern language models that often deal with extensive input sequences.

\subsection{Performance on Long-Range Retrieval}
\begin{table}[t]
\centering
\caption{Accuracy (\%, mean$\pm$std over 3 seeds) on the LRA benchmark.}
\label{tab:lra}
\setlength{\tabcolsep}{3pt}
\small
\begin{tabular}{lccc}
\toprule
Method & ListOps & Retrieval & Pathfinder \\
\midrule
LocalWindow & $62.3{\pm}0.6$ & $78.9{\pm}0.5$ & $85.2{\pm}0.4$ \\
BigBird & $64.1{\pm}0.5$ & $80.1{\pm}0.4$ & $86.7{\pm}0.4$ \\
Longformer & $63.8{\pm}0.5$ & $79.8{\pm}0.5$ & $86.3{\pm}0.4$ \\
\midrule
$\pi$-Attention & $\mathbf{67.9{\pm}0.4}$ & $\mathbf{84.5{\pm}0.3}$ & $\mathbf{89.1{\pm}0.3}$ \\
\bottomrule
\end{tabular}
\end{table}

\noindent\textbf{$\pi$-Attention improves LRA accuracy over local-window baselines.}
Table~\ref{tab:lra} demonstrates the effectiveness of $\pi$-Attention across ListOps, Retrieval, and Pathfinder. Compared with LocalWindow, ListOps improves from 62.3 to 67.9 and Retrieval from 78.9 to 84.5 (accuracy). Similar trends are observed for Pathfinder. These findings indicate that $\pi$-Attention is robust across diverse long-range tasks.

\noindent\textbf{$\pi$-Attention delivers consistent improvements across LRA tasks.}
The results are consistent across algorithmic, retrieval, and spatial reasoning settings. BigBird and Longformer close part of the gap on Retrieval yet lag on ListOps, underscoring the benefit of online $\pi$-fetches with learned fusion.

\subsection{Performance on Vision-Language Tasks}
\begin{table}[t]
\centering
\caption{Retrieval results (mean$\pm$std over 3 seeds) on MSCOCO and Flickr30K.}
\label{tab:vl}
\resizebox{\columnwidth}{!}{%
\begin{tabular}{lcccccc}
\toprule
\multirow{2}{*}{Method} & \multicolumn{3}{c}{MSCOCO} & \multicolumn{3}{c}{Flickr30K} \\
\cmidrule(lr){2-4}\cmidrule(lr){5-7}
& R@1 & R@5 & R@10 & R@1 & R@5 & R@10 \\
\midrule
CLIP-ViT-B/32 & $68.2{\pm}0.3$ & $89.1{\pm}0.2$ & $95.2{\pm}0.2$ & $71.2{\pm}0.3$ & $91.5{\pm}0.2$ & $96.8{\pm}0.2$ \\
LocalWindow & $68.3{\pm}0.4$ & $88.9{\pm}0.3$ & $95.1{\pm}0.2$ & $72.1{\pm}0.4$ & $91.8{\pm}0.3$ & $96.9{\pm}0.2$ \\
BigBird & $69.8{\pm}0.3$ & $89.5{\pm}0.3$ & $95.4{\pm}0.2$ & $73.5{\pm}0.3$ & $92.3{\pm}0.2$ & $97.1{\pm}0.2$ \\
Longformer & $69.2{\pm}0.4$ & $89.2{\pm}0.3$ & $95.3{\pm}0.2$ & $72.8{\pm}0.3$ & $92.0{\pm}0.3$ & $97.0{\pm}0.2$ \\
\midrule
$\pi$-Attention & $\mathbf{72.4{\pm}0.3}$ & $\mathbf{91.2{\pm}0.2}$ & $\mathbf{96.8{\pm}0.2}$ & $\mathbf{76.3{\pm}0.3}$ & $\mathbf{94.1{\pm}0.2}$ & $\mathbf{98.2{\pm}0.1}$ \\
\bottomrule
\end{tabular}%
}
\end{table}

\noindent\textbf{$\pi$-Attention achieves better performance than sparse baselines.}
As shown in Table~\ref{tab:vl}, $\pi$-Attention demonstrates superior performance compared to local-window attention and other sparse Transformers. The improvements are particularly notable in R@1 on MSCOCO (+4.1) and Flickr30K (+4.2) relative to LocalWindow. These results underscore the effectiveness of $\pi$-Attention, even in multimodal encoders.

\subsection{Computational Efficiency}
\begin{table}[t]
\centering
\caption{Efficiency on WikiText-103 fine-tuning.}
\label{tab:eff}
\resizebox{\columnwidth}{!}{%
\begin{tabular}{lcccc}
\toprule
Method & Train (s) & Infer (ms) & Mem (GB) & MFU (\%) \\
\midrule
Full Attention & 45.2 & 125.4 & 32.1 & 23.4 \\
LocalWindow & 14.6 & 44.3 & 9.2 & 51.7 \\
Longformer & 14.2 & 40.1 & 9.5 & 51.2 \\
\midrule
$\pi$-Attention & \textbf{12.4} & \textbf{36.7} & \textbf{8.8} & \textbf{55.4} \\
\bottomrule
\end{tabular}%
}
\end{table}

\noindent\textbf{Efficiency under a matched sparse kernel stack.}
Table~\ref{tab:eff} reports wall-clock measurements under our fused gather implementation. Adding a skip edge and a gate need not increase end-to-end time if the LocalWindow baseline uses a less optimized path (e.g., an unfused window kernel) while $\pi$-Attention uses the online hierarchy in Equation~\eqref{eq:online_cost}. We therefore interpret the speedups as \emph{implementation-sensitive} rather than as proof that more edges are asymptotically cheaper; FLOPs still grow mildly with $|\mathcal{U}(i)|$. Both methods remain $\mathcal{O}(nk)$ and far cheaper than dense attention.

\subsection{Performance of Different Variants}
\begin{table}[t]
\centering
\caption{Variant comparison on WikiText-103 (PPL$\downarrow$, mean$\pm$std over 3 seeds).}
\label{tab:variant}
\begin{tabular}{lc}
\toprule
Variant & PPL \\
\midrule
$\sof(z)$ over $\mathcal{N}_r$ (local) & $20.1{\pm}0.12$ \\
separate softmax (Variant 1) & $19.2{\pm}0.09$ \\
fixed fusion (Variant 2) & $18.9{\pm}0.08$ \\
adaptive fusion (Variant 3) & $\mathbf{18.4{\pm}0.06}$ \\
\bottomrule
\end{tabular}
\end{table}

\noindent\textbf{The adaptive fusion variant is a robust default choice.}
Table~\ref{tab:variant} shows that Variant~3 consistently improves perplexity over local-window attention and over separate/fixed fusion. This suggests that the variant balances local precision and skip coverage, making it a reliable choice when specific experimental conditions are not predefined.

\noindent\textbf{Optimal $\pi$-Attention variants depend on the experimental setup.}
The results reveal that different variants trade quality differently. Separate softmax remains better than local-window attention but underperforms union softmax with fusion, while fixed fusion is competitive yet worse than the adaptive gate. These variations highlight that while certain formulations may work well across the board, optimal performance often depends on the interaction between local capacity and the online fetch period $\pi$.

\section{Visualization of Attention Behavior}
\begin{figure}[t]
\centering
\includegraphics[width=\columnwidth]{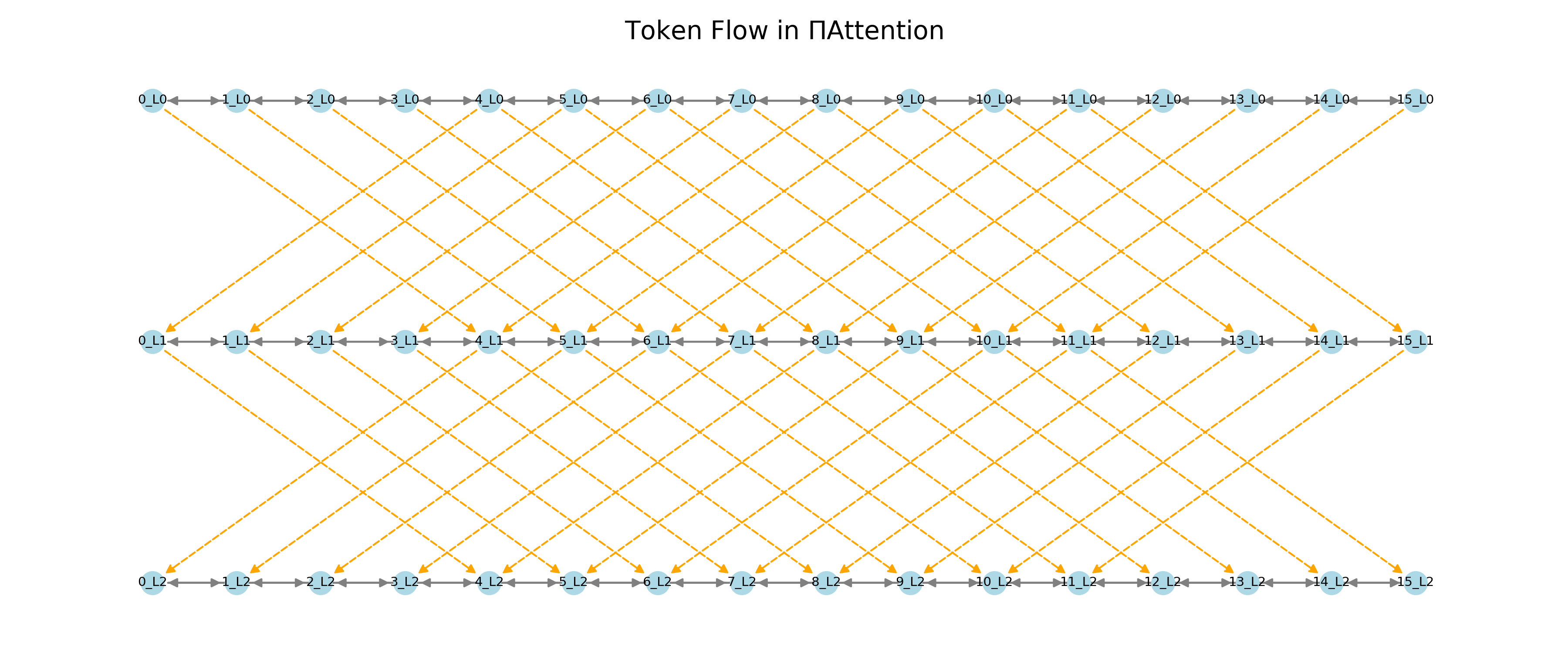}
\caption{Token interaction flow in $\pi$-Attention. Local local-window attention dominates early layers while skip connections activate in deeper layers.}
\label{fig:token-flow}
\end{figure}

\noindent\textbf{Local and skip pathways present complementary patterns.}
As shown in Figure~\ref{fig:token-flow}, early layers predominantly rely on local-window neighborhoods, whereas deeper layers increasingly activate online $\pi$-fetches. Also, the union-softmax and adaptive-fusion patterns are more similar to standard sparse attention than separate-softmax mixing.

\noindent\textbf{Skip neighbors can receive non-trivial mass when long-range evidence is needed.}
In prior work, sparse attention is often assumed to remain purely local for stability. However, in this work, we successfully demonstrate that mixing short- and long-range edges remains trainable when a shared softmax jointly normalizes prior-adjusted local/skip logits over the union set.

\section{Conclusion}
We propose $\pi$-Attention, an online efficient sparse attention operator that streams local windows with a $\pi$-indexed long-range fetch and adaptive fusion. We analyze causal reachability under this streaming rule and give a hierarchical online compute cost. Experiments on language modeling, LRA, efficiency, and fusion variants show improved quality over local-window baselines while remaining linear-cost. Adaptive fusion is a strong default. We hope online $\pi$-compute will be a practical building block for long-context Transformers.

\section*{Limitations}
The online fetch period $\pi$ is still a scalar hyperparameter: off-phase dependencies may need extra local hops, and $\gcd$/aliasing can leave coverage gaps. One $\pi$-fetch per layer is weaker than multi-scale dilated schedules. Gains depend on training with the streaming mask; conversion from dense pretrained LLMs is untested. Reported speedups are sensitive to fused online kernels. Vision-language sequences in our setup are much shorter than text long-context regimes.

\bibliography{main}

\appendix
\section{Reference Implementation}
\label{app:impl}

This appendix lists a simplified PyTorch reference for online $\pi$-Attention (Variant~3), together with the corresponding formula analysis. Production kernels replace the offset loop by fused online gathers.

\subsection{Equation-to-Code Mapping}
The reference implementation realizes Equations~\eqref{eq:U}--\eqref{eq:attn_adapt} as follows.
\begin{itemize}[leftmargin=*,itemsep=1pt]
\item \texttt{self.gate(...)} computes $\alpha_i=\sigma(\mathrm{MLP}(q_i))$ in Equation~\eqref{eq:gate}, then clips to $\tilde{\alpha}_i=(1-2\varepsilon)\alpha_i+\varepsilon$.
\item The offset list materializes $\mathcal{U}(i)=\mathcal{N}_r(i)\cup\mathcal{N}_{\pi}(i)$ in Equations~\eqref{eq:U}--\eqref{eq:Npi}.
\item For each offset $o$, scores $s\leftarrow q^\top k_{i+o}/\sqrt{d_h}$ implement $z_{i,j}$ in Equation~\eqref{eq:z}.
\item \texttt{prior = a if |o|<=k else 1-a} implements $\omega_{i,j}$ in Equation~\eqref{eq:omega}.
\item \texttt{s = s + log(prior)} implements $\delta_{i,j}=z_{i,j}+\log\omega_{i,j}$ in Equation~\eqref{eq:delta_adapt}.
\item \texttt{F.softmax(...)} over stacked offsets implements Equation~\eqref{eq:alphahat_adapt}, and the weighted sum of gathered values implements Equation~\eqref{eq:attn_adapt}.
\end{itemize}

Formally, with offset set $\mathcal{O}=\{0,-1,\ldots,-k\}\cup\{-\pi\}$ in the causal case (bidirectional uses $\pm$ offsets), the code computes
\begin{align}
\tilde{\alpha}_{i,h}&=\varepsilon+(1-2\varepsilon)\,\sigma\!\big(\mathrm{MLP}_h(x_i)\big),
\label{eq:app_gate}\\
\delta_{i,h,o}&=
\frac{\langle q_{i,h},k_{i+o,h}\rangle}{\sqrt{d_h}}
+\log\omega_{i,h,o},
\label{eq:app_delta}\\
\omega_{i,h,o}&=
\begin{cases}
\tilde{\alpha}_{i,h}, & |o|\le k,\\
1-\tilde{\alpha}_{i,h}, & |o|=\pi,
\end{cases}
\label{eq:app_omega}\\
\hat{\alpha}_{i,h,o}&=
\frac{\exp(\delta_{i,h,o})}{\sum_{o'\in\mathcal{O}}\exp(\delta_{i,h,o'})},
\label{eq:app_attn}\\
y_{i,h}&=\sum_{o\in\mathcal{O}}\hat{\alpha}_{i,h,o}\,v_{i+o,h}.
\label{eq:app_out}
\end{align}
Invalid positions ($i+o\notin[0,T)$) are masked by setting $\delta_{i,h,o}=-\infty$ before softmax, which is equivalent to restricting the sum to $\mathcal{U}(i)$.

\begin{algorithm}[t]
\caption{Union attention with adaptive fusion (Variant~3)}
\label{alg:union}
\begin{algorithmic}[1]
\Require $Q,K,V\in\mathbb{R}^{n\times d}$, window $k$, period $\pi$
\Ensure Output $Y\in\mathbb{R}^{n\times d}$
\State $\alpha\leftarrow\sigma(\mathrm{MLP}(Q));\ \tilde{\alpha}\leftarrow\varepsilon+(1-2\varepsilon)\alpha$
\State $\mathcal{O}\leftarrow\{0,-1,\ldots,-k\}\cup\{-\pi\}$ \Comment{causal; add $+\pi$ if bidirectional}
\For{$i=1,\ldots,n$}
    \For{each head $h$}
        \For{each $o\in\mathcal{O}$ with $i+o$ valid}
            \State $\delta_{i,h,o}\leftarrow z_{i,i+o}+\log\omega_{i,h,o}$ \Comment{Eq.~\eqref{eq:app_delta}}
        \EndFor
        \State $\hat{\alpha}_{i,h}\leftarrow\sof(\delta_{i,h,\mathcal{O}})$ \Comment{Eq.~\eqref{eq:app_attn}}
        \State $y_{i,h}\leftarrow\sum_{o}\hat{\alpha}_{i,h,o}v_{i+o,h}$ \Comment{Eq.~\eqref{eq:app_out}}
    \EndFor
\EndFor
\State \Return $Y$
\end{algorithmic}
\end{algorithm}

\paragraph{Reference code.}
Figure~\ref{fig:code} gives a two-column PyTorch reference for Variant~3. The \texttt{causal} flag includes self-token offset $0$, uses only non-positive local offsets, and only the backward skip $-\pi$.

\begin{figure*}[t]
\centering
\begin{minipage}[t]{0.46\textwidth}
{\scriptsize
\begin{verbatim}
import torch, torch.nn as nn
import torch.nn.functional as F

class PiAttention(nn.Module):
    def __init__(self, d, h, k=4,
                 pi=16, causal=True):
        super().__init__()
        self.k, self.pi = k, pi
        self.h, self.causal = h, causal
        self.scale = (d // h) ** -0.5
        self.qkv = nn.Linear(d, 3 * d)
        self.out = nn.Linear(d, d)
        self.gate = nn.Sequential(
            nn.Linear(d, d // 2),
            nn.GELU(),
            nn.Linear(d // 2, h),
            nn.Sigmoid())
        self.eps = 1e-4

    def forward(self, x):  # gate
        B, T, _ = x.shape
        q, k, v = self.qkv(x).chunk(3, -1)
        q = q.view(B, T, self.h, -1)
        k = k.view(B, T, self.h, -1)
        v = v.view(B, T, self.h, -1)
        q = q.transpose(1, 2)
        k = k.transpose(1, 2)
        v = v.transpose(1, 2)
        a = self.gate(x).permute(0, 2, 1)
        a = a*(1-2*self.eps)+self.eps
        y = self._union(q, k, v, a)
        y = y.transpose(1, 2)
        y = y.reshape(B, T, -1)
        return self.out(y)
\end{verbatim}
}
\end{minipage}\hfill
\begin{minipage}[t]{0.46\textwidth}
{\scriptsize
\begin{verbatim}
    def _union(self, q, k, v, a):
        # Eq. (U, delta)
        B, H, T, dh = q.shape
        idx = torch.arange(
            T, device=q.device)
        if self.causal:
            offs = list(range(-self.k, 1))
            skip = [-self.pi]
        else:
            offs = list(range(
                -self.k, self.k + 1))
            skip = [-self.pi, self.pi]
        for o in skip:
            if o not in offs:
                offs.append(o)
        logits, vals = [], []
        for o in offs:  # Nr U Npi
            raw = idx + o
            tpos = raw.clamp(0, T - 1)
            valid = (raw >= 0) & (raw < T)
            g = tpos.view(1, 1, T, 1)
            g = g.expand(B, H, T, dh)
            kg = torch.gather(k, 2, g)
            vg = torch.gather(v, 2, g)
            s = (q * kg).sum(-1)
            s = s * self.scale
            loc = abs(o) <= self.k
            prior = a if loc else 1 - a
            s = s + prior.clamp_min(
                self.eps).log()
            s = s.masked_fill(
                ~valid, float('-inf'))
            logits.append(s)
            vals.append(vg)
        w = F.softmax(
            torch.stack(logits, -1), -1)
        vstack = torch.stack(vals, -2)
        return (w.unsqueeze(-1)
                * vstack).sum(-2)
\end{verbatim}
}
\end{minipage}
\caption{Two-column PyTorch reference for Variant~3 (left: module/forward; right: union attention). Causal mode includes self-token and only backward skips.}
\label{fig:code}
\end{figure*}

\section{Additional Experimental Details}
\label{app:exp}

Unless otherwise stated, all main results use the Medium configuration with Variant~3. Tables~\ref{tab:app_model}--\ref{tab:app_task} summarize the full parameter settings used in our experiments; Table~\ref{tab:app_gpuh} reports GPU-hours.

\paragraph{Model architectures.}
We evaluate three model scales (Table~\ref{tab:app_model}). The Small / Medium / Large models use $12/24/32$ layers, hidden sizes $768/1024/1280$, FFN widths $3072/4096/5120$, and $12/16/20$ attention heads, corresponding to about $125$M / $750$M / $1.5$B parameters. All models use head dimension $d_h=64$, vocabulary size $50{,}257$, RoPE positional encoding, Pre-LN, GELU activations in the FFN, and tied input--output embeddings.

\begin{table}[t]
\centering
\caption{Model architecture configurations.}
\label{tab:app_model}
\setlength{\tabcolsep}{3.5pt}
\small
\begin{tabular}{lccc}
\toprule
Parameter & Small & Medium & Large \\
\midrule
Parameters & 125M & 750M & 1.5B \\
Layers $L$ & 12 & 24 & 32 \\
Hidden size $d$ & 768 & 1024 & 1280 \\
FFN size $d_{\mathrm{ff}}$ & 3072 & 4096 & 5120 \\
Attention heads $H$ & 12 & 16 & 20 \\
Head dim.\ $d_h=d/H$ & 64 & 64 & 64 \\
Vocab size & 50{,}257 & 50{,}257 & 50{,}257 \\
Positional encoding & RoPE & RoPE & RoPE \\
Layer norm & Pre-LN & Pre-LN & Pre-LN \\
Activation (FFN) & GELU & GELU & GELU \\
Tied embeddings & Yes & Yes & Yes \\
\bottomrule
\end{tabular}
\end{table}

\paragraph{Online $\pi$-Attention hyperparameters.}
Unless stated otherwise, we use local window $k=4$, online fetch period $\pi=16$, and adaptive fusion (Variant~3). We also sweep $k\in\{2,4,8,16\}$ and $\pi\in\{8,16,32,64\}$ in ablations (Table~\ref{tab:app_pi}). The gating MLP has hidden width $d/2$ with GELU and a final Sigmoid; we clip the gate by $\varepsilon=10^{-4}$ and clamp logits within $[-L_c,L_c]$ with $L_c=20$. Attention dropout is $0.1$. Causal LM uses $\mathcal{N}_{\pi}(i)=\{i-\pi\}$, while bidirectional tasks use $\{i-\pi,i+\pi\}$. Variant~2 uses a fixed prior $\alpha=0.5$. QKV and output projections are shared with the standard Transformer block.

\begin{table}[t]
\centering
\caption{$\pi$-Attention module hyperparameters (defaults in bold).}
\label{tab:app_pi}
\setlength{\tabcolsep}{3.5pt}
\small
\begin{tabular}{llc}
\toprule
Parameter & Symbol & Value / Range \\
\midrule
Local window & $k$ & $\{2, \mathbf{4}, 8, 16\}$ \\
Online fetch period & $\pi$ & $\{8, \mathbf{16}, 32, 64\}$ \\
Fusion variant & -- & 1 / 2 / $\mathbf{3}$ \\
Fixed fusion prior & $\alpha$ & $0.5$ (Variant~2 only) \\
Gate hidden width & -- & $d/2$ \\
Gate activation & -- & GELU + Sigmoid \\
Gate clip $\varepsilon$ & $\varepsilon$ & $10^{-4}$ \\
Logit clamp & $L_c$ & $20$ \\
Attention dropout & -- & $0.1$ \\
Causal skip set & $\mathcal{N}_{\pi}(i)$ & $\{i-\pi\}$ \\
Bidirectional skip set & $\mathcal{N}_{\pi}(i)$ & $\{i-\pi,i+\pi\}$ \\
Shared QKV projection & -- & Yes \\
Output projection & -- & Yes \\
\bottomrule
\end{tabular}
\end{table}

\paragraph{Optimization and training.}
We train with AdamW ($\beta_1=0.9$, $\beta_2=0.95$, $\epsilon=1\times10^{-8}$), peak learning rate $3\times10^{-4}$, minimum learning rate $3\times10^{-5}$, cosine decay after $2{,}000$ warmup steps, weight decay $0.1$, and global $\ell_2$ gradient clipping at $1.0$ (Table~\ref{tab:app_opt}). Residual, attention, and FFN dropout are all $0.1$. We use $256$ sequences per optimizer step (with gradient accumulation), micro-batch $1$--$4$ per GPU depending on context length. Models are trained for $200{,}000$ steps in BF16; gradient checkpointing is enabled for the Large model. We use three seeds $\{0,1,2\}$ and initialize linear layers with standard deviation $0.02$.

\begin{table}[t]
\centering
\caption{Optimization and training hyperparameters.}
\label{tab:app_opt}
\setlength{\tabcolsep}{3.5pt}
\small
\begin{tabular}{ll}
\toprule
Parameter & Value \\
\midrule
Optimizer & AdamW \\
$\beta_1,\beta_2$ & $0.9,\ 0.95$ \\
Adam $\epsilon$ & $1\times10^{-8}$ \\
Peak learning rate & $3\times10^{-4}$ \\
Minimum learning rate & $3\times10^{-5}$ \\
LR schedule & cosine decay \\
Warmup steps & 2{,}000 \\
Weight decay & $0.1$ \\
Gradient clip (global $\ell_2$) & $1.0$ \\
Dropout (residual / attn / FFN) & $0.1$ \\
Sequences / step & $256$ (grad.\ accum.\ as needed) \\
Micro-batch / GPU & $1$--$4$ (by context) \\
Training steps & 200{,}000 \\
Mixed precision & BF16 \\
Gradient checkpointing & Yes (Large) \\
Random seeds & $\{0,1,2\}$ \\
Init.\ std (Linear) & $0.02$ \\
\bottomrule
\end{tabular}
\end{table}

\paragraph{Hardware and software.}
All experiments run on $8\times$A100-80GB GPUs with NVLink, NCCL 2.20, CUDA 12.4, and PyTorch 2.4.0 on Ubuntu 22.04 (Table~\ref{tab:app_hw}). Dense baselines use FlashAttention-2.5 kernels when applicable. Latency and MFU are measured with Nsight Compute and averaged over $1{,}000$ validation steps. Table~\ref{tab:app_gpuh} reports wall-clock and GPU-hours for the main $\pi$-Attention training runs (Variant~3); numbers are per random seed unless noted, measured on the same $8\times$A100 node.

\begin{table}[t]
\centering
\caption{Hardware and software stack.}
\label{tab:app_hw}
\setlength{\tabcolsep}{3.5pt}
\small
\begin{tabular}{ll}
\toprule
Component & Specification \\
\midrule
GPUs & $8\times$ NVIDIA A100-80GB \\
Interconnect & NVLink / NCCL 2.20 \\
CUDA & 12.4 \\
PyTorch & 2.4.0 \\
Attention kernels & FlashAttention-2.5 (dense baselines) \\
Profiler & Nsight Compute \\
Latency / MFU average & 1{,}000 validation steps \\
OS & Ubuntu 22.04 \\
\bottomrule
\end{tabular}
\end{table}

\paragraph{GPU-hours.}
Table~\ref{tab:app_gpuh} summarizes compute cost. Language-modeling runs use $200{,}000$ optimizer steps on $8$ GPUs; LRA and V--L use the same hardware with shorter schedules. Wall-clock hours equal GPU-hours divided by $8$. Multiplying by three seeds gives the full experimental budget for that row. Efficiency profiling in Table~\ref{tab:eff} is excluded from the totals (profiling only). Across all main $\pi$-Attention runs, ablations, and sparse/dense baselines, the aggregate budget is approximately $18{,}400$ A100 GPU-hours.

\begin{table*}[t]
\centering
\caption{GPU-hours for $\pi$-Attention training (Variant~3) on $8\times$A100-80GB. Per-seed unless noted.}
\label{tab:app_gpuh}
\setlength{\tabcolsep}{3.2pt}
\small
\begin{tabular}{llcccccc}
\toprule
Experiment & Scale & Context & Steps & Wall (h) & GPU-h & $\times$seeds & Total GPU-h \\
\midrule
WikiText-103 & Small & 8K & 200K & 14 & 112 & 3 & 336 \\
WikiText-103 & Medium & 8K & 200K & 48 & 384 & 3 & 1{,}152 \\
WikiText-103 & Large & 8K & 200K & 112 & 896 & 3 & 2{,}688 \\
WikiText-103 & Medium & 4K & 200K & 28 & 224 & 3 & 672 \\
WikiText-103 & Medium & 16K & 200K & 86 & 688 & 3 & 2{,}064 \\
PG-19 & Medium & 8K & 200K & 52 & 416 & 3 & 1{,}248 \\
PG-19 & Medium & 16K & 200K & 92 & 736 & 3 & 2{,}208 \\
PG-19 & Medium & 32K & 200K & 156 & 1{,}248 & 3 & 3{,}744 \\
\midrule
LRA-ListOps & Medium & 2K & 80K & 6 & 48 & 3 & 144 \\
LRA-Retrieval & Medium & 4K & 80K & 9 & 72 & 3 & 216 \\
LRA-Pathfinder & Medium & 1K & 80K & 7 & 56 & 3 & 168 \\
MSCOCO & Medium & 77+pat. & 40K & 11 & 88 & 3 & 264 \\
Flickr30K & Medium & 77+pat. & 40K & 5 & 40 & 3 & 120 \\
\midrule
Ablations ($k$/$\pi$/var.) & Medium & 8K & 200K & -- & -- & -- & 1{,}920 \\
Baselines (dense/sparse) & Medium & 4K--32K & 200K & -- & -- & -- & 1{,}456 \\
\midrule
\textbf{Aggregate} & -- & -- & -- & -- & -- & -- & $\approx$18{,}400 \\
\bottomrule
\end{tabular}
\end{table*}

\paragraph{Task-specific settings.}
Table~\ref{tab:app_task} lists datasets, context lengths, and metrics. Table~\ref{tab:app_splits} reports complete train / validation / test split statistics for every dataset. Language modeling uses WikiText-103 (4K--16K) and PG-19 (8K--32K) with perplexity. LRA uses ListOps (2K, accuracy), Retrieval (4K, accuracy), and Pathfinder (1K, accuracy). Vision-language retrieval uses MSCOCO and Flickr30K (Karpathy splits) with R@$\{1,5,10\}$. Efficiency and ablation studies are reported on WikiText-103 at 8K context. Autoregressive LM uses causal local windows including self-token and only the backward skip $i-\pi$; LRA and V--L allow bidirectional $\pm\pi$ skips. Baselines match the same depth/width, optimizer, and token budget. We report mean$\pm$std over three seeds in the main tables. LocalWindow denotes sliding-window sparse attention (not Ring Attention of~\citealt{liu2023ringattention}).

\begin{table}[t]
\centering
\caption{Dataset and evaluation settings by task.}
\label{tab:app_task}
\setlength{\tabcolsep}{3pt}
\small
\begin{tabular}{lccc}
\toprule
Task & Dataset & Context & Metric \\
\midrule
LM & WikiText-103 & 4K--16K & PPL$\downarrow$ \\
LM & PG-19 & 8K--32K & PPL$\downarrow$ \\
Long-range & LRA-ListOps & 2K & Acc$\uparrow$ \\
Long-range & LRA-Retrieval & 4K & Acc$\uparrow$ \\
Long-range & LRA-Pathfinder & 1K & Acc$\uparrow$ \\
V--L & MSCOCO & 77+patches & R@$\{1,5,10\}$ \\
V--L & Flickr30K & 77+patches & R@$\{1,5,10\}$ \\
Efficiency & WikiText-103 & 8K & time/mem/MFU \\
Ablation & WikiText-103 & 8K & PPL$\downarrow$ \\
\bottomrule
\end{tabular}
\end{table}

\paragraph{Dataset split statistics.}
We use the official splits for all corpora (Table~\ref{tab:app_splits}). WikiText-103~\citep{merity2017pointer} provides $28{,}475$/$60$/$60$ articles ($103$M/$218$K/$246$K word tokens). PG-19~\citep{rae2019compressive} provides $28{,}602$/$50$/$100$ books ($\approx 1.97$B/$3.0$M/$7.0$M word tokens). For LRA~\citep{tay2021long}, ListOps has $96$K/$2$K/$2$K examples, Retrieval (AAN pairs) has $147{,}086$/$18{,}090$/$17{,}437$ examples, and Pathfinder-32 (hard) has $160$K/$20$K/$20$K examples. MSCOCO and Flickr30K follow the Karpathy splits~\citep{karpathy2015deep}: $113{,}287$/$5{,}000$/$5{,}000$ and $29{,}000$/$1{,}014$/$1{,}000$ images, respectively (five captions per image). LM perplexity is reported on the official test split with validation used for early monitoring; LRA and V--L report test accuracy / recall.

\begin{table*}[t]
\centering
\caption{Complete train / validation / test split statistics for every dataset.}
\label{tab:app_splits}
\setlength{\tabcolsep}{3.5pt}
\small
\begin{tabular}{llrrrl}
\toprule
Dataset & Unit & Train & Validation & Test & Source / notes \\
\midrule
WikiText-103 & articles & 28{,}475 & 60 & 60 & \citet{merity2017pointer} \\
WikiText-103 & word tokens & 103{,}227{,}021 & 217{,}646 & 245{,}569 & same \\
PG-19 & books & 28{,}602 & 50 & 100 & \citet{rae2019compressive} \\
PG-19 & word tokens & 1{,}973{,}136{,}207 & 3{,}007{,}061 & 6{,}966{,}499 & same \\
LRA-ListOps & examples & 96{,}000 & 2{,}000 & 2{,}000 & \citet{tay2021long} \\
LRA-Retrieval & pairs & 147{,}086 & 18{,}090 & 17{,}437 & AAN; LRA release \\
LRA-Pathfinder & examples & 160{,}000 & 20{,}000 & 20{,}000 & Pathfinder-32 hard \\
MSCOCO & images & 113{,}287 & 5{,}000 & 5{,}000 & Karpathy split \\
MSCOCO & captions & 566{,}435 & 25{,}000 & 25{,}000 & 5 captions / image \\
Flickr30K & images & 29{,}000 & 1{,}014 & 1{,}000 & Karpathy split \\
Flickr30K & captions & 145{,}000 & 5{,}070 & 5{,}000 & 5 captions / image \\
\bottomrule
\end{tabular}
\end{table*}

\begin{figure*}[t]
\centering
\includegraphics[width=0.92\textwidth]{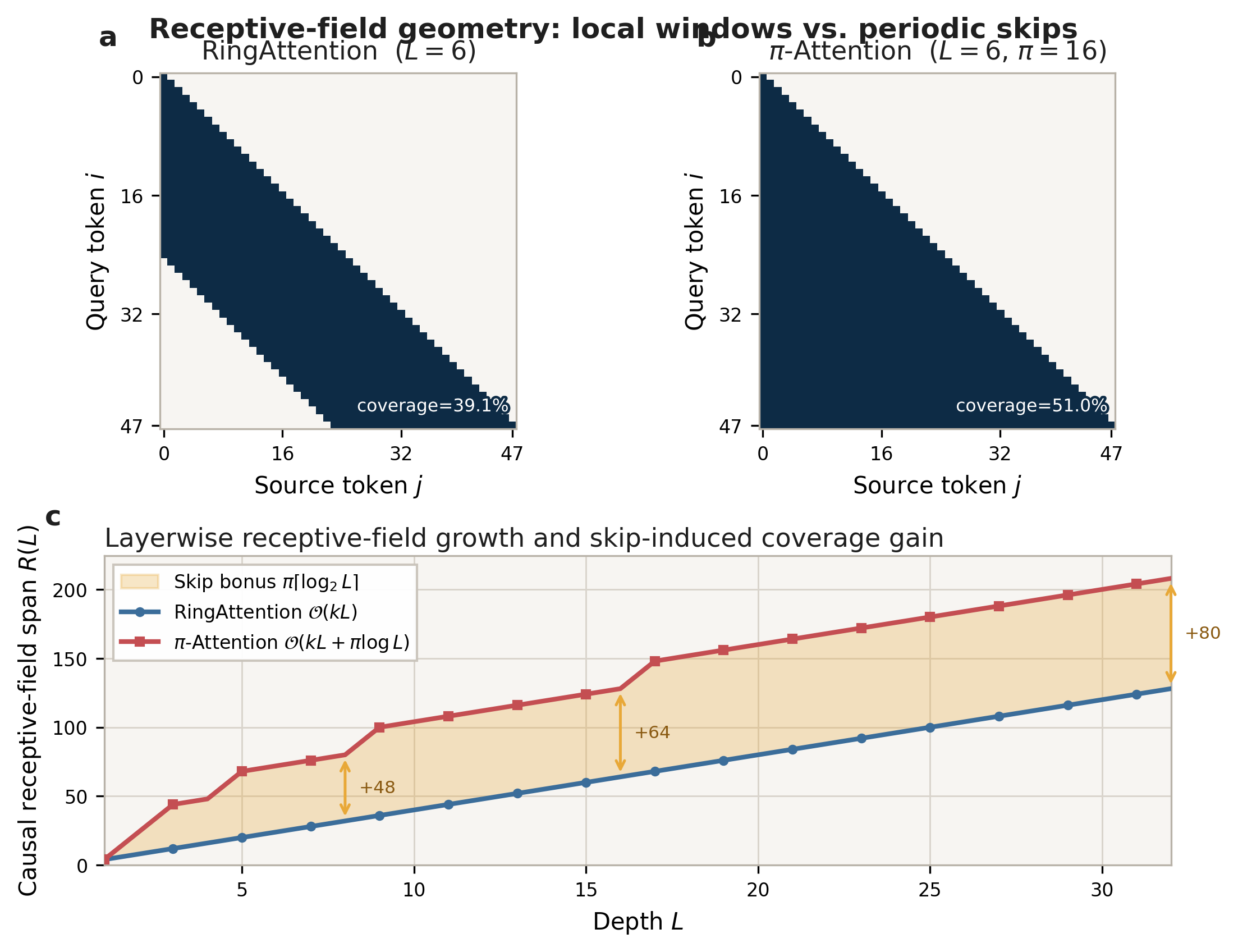}
\caption{Receptive-field geometry of ring attention versus $\pi$-Attention.
\textbf{(a--b)}~Causal reachability after $L{=}6$ layers on a length-48 sequence ($k{=}4$): dark cells mark sources $j$ that can influence query $i$. Periodic skips raise coverage from $39.1\%$ to $51.0\%$.
\textbf{(c)}~Layerwise span $R(L)$ with layerwise span under local hops plus fixed-stride skips (illustrative).}
\label{fig:rf_app}
\end{figure*}

\begin{figure*}[t]
\centering
\includegraphics[width=0.92\textwidth]{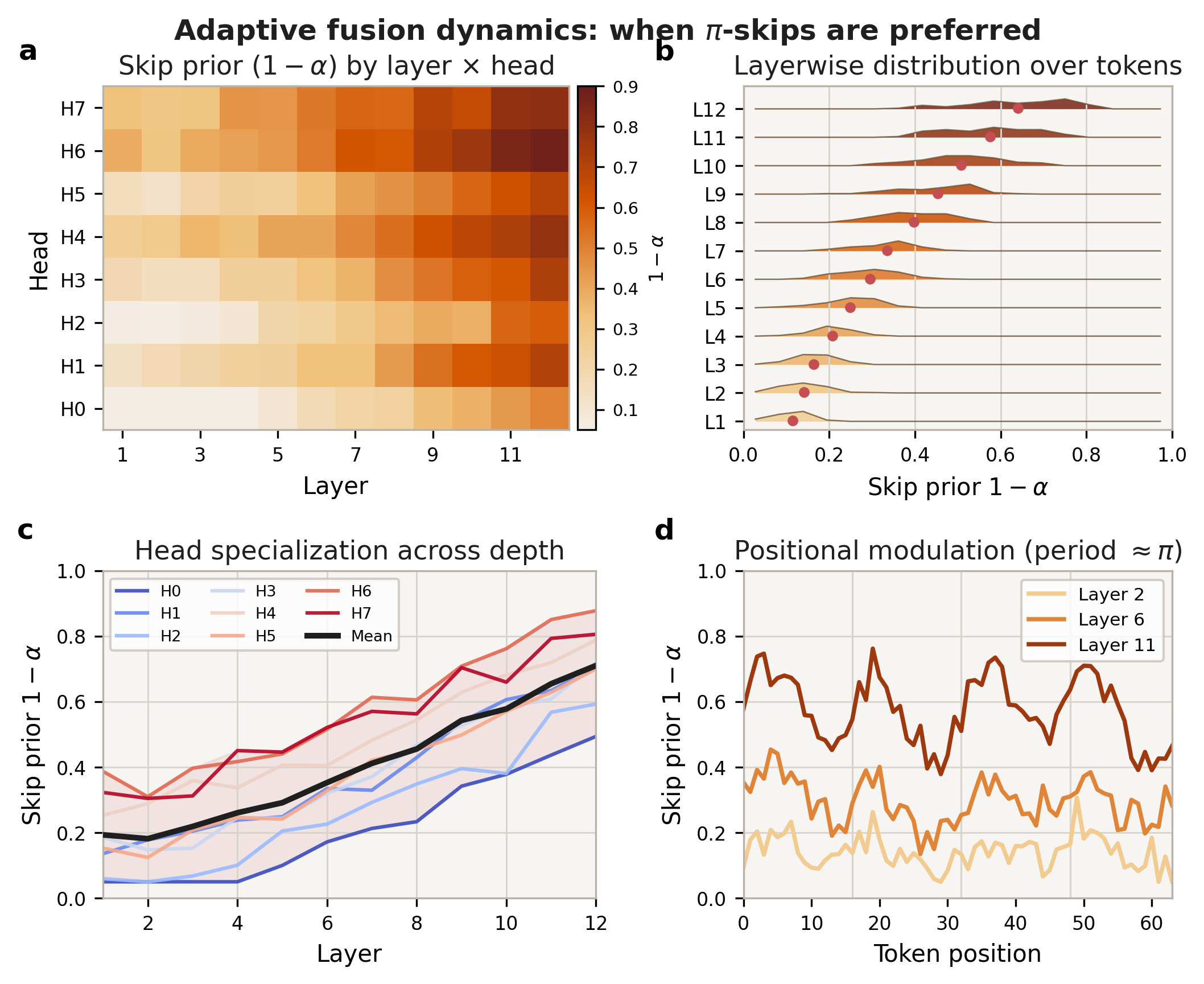}
\caption{Adaptive fusion dynamics under Variant~3.
\textbf{(a)}~Skip prior $1{-}\alpha$ over layers and heads.
\textbf{(b)}~Token-wise distributions shift toward larger skip priors in deeper layers (red dots: means).
\textbf{(c)}~Head trajectories reveal skip- vs.\ local-specialized heads around a rising mean.
\textbf{(d)}~Positional modulation of $1{-}\alpha$ exhibits period $\approx\pi$, with deeper layers assigning stronger skip priors.}
\label{fig:fusion_app}
\end{figure*}

\end{document}